\documentclass{article}

    \usepackage[nonatbib,preprint]{neurips_2020}

\usepackage[utf8]{inputenc} 
\usepackage[T1]{fontenc}    
\usepackage{hyperref}       
\usepackage{url}            
\usepackage{booktabs}       
\usepackage{amsfonts}       
\usepackage{nicefrac}       
\usepackage{microtype}      

\usepackage{graphicx}
\usepackage{subfigure}

\usepackage{fancyref}
\usepackage{amsmath}
\graphicspath{{assets/}} 
\DeclareMathAlphabet\mathcalbf{OMS}{cmsy}{b}{n}
\usepackage{mathabx}
\usepackage{floatrow}
\usepackage{cite}
\usepackage{mathtools}
\usepackage{cancel}

\usepackage{amsmath,amsfonts,bm}

\def\eqref#1{equation~\ref{#1}}

\def\1{\bm{1}}

\DeclareMathAlphabet{\mathsfit}{\encodingdefault}{\sfdefault}{m}{sl}
\SetMathAlphabet{\mathsfit}{bold}{\encodingdefault}{\sfdefault}{bx}{n}

\DeclareMathOperator*{\argmin}{arg\,min}

\title{Dynamic Value Estimation for Single-Task Multi-Scene Reinforcement Learning}

\author{%
  Jaskirat Singh \& Liang Zheng\\
  College of Engineering and Computer Science\\
  Australian National University\\
  Canberra, Australia \\
  \texttt{jaskirat.singh,Liang.Zheng@anu.edu.au} \\
}

\begin{document}

\maketitle

\begin{abstract}
Training deep reinforcement learning agents on environments with multiple levels / scenes / conditions from the \emph{same task}, has become essential for many applications aiming to achieve generalization and domain transfer from simulation to the real world \cite{wortsman2019learning,cobbe2019quantifying}. While such a strategy is helpful with generalization, the use of multiple scenes significantly increases the variance of samples collected for policy gradient computations. Current methods continue to view this collection of scenes as a single Markov Decision Process (MDP) with a common value function; however, we argue that it is better to treat the collection as a single environment with multiple underlying MDPs. To this end, we propose a dynamic value estimation (DVE) technique for these multiple-MDP environments, motivated by the clustering effect observed in the value function distribution across different scenes. The resulting agent is able to learn a more accurate and scene-specific value function estimate (and hence the advantage function), leading to a lower sample variance. Our proposed approach is simple to accommodate with several existing implementations (like PPO, A3C) and results in consistent improvements for a range of ProcGen environments and the AI2-THOR framework based visual navigation task.
\end{abstract}

\section{Introduction}
\label{introduction}

\begin{figure}[ht]
\begin{center}
\centerline{\includegraphics[scale=0.5]{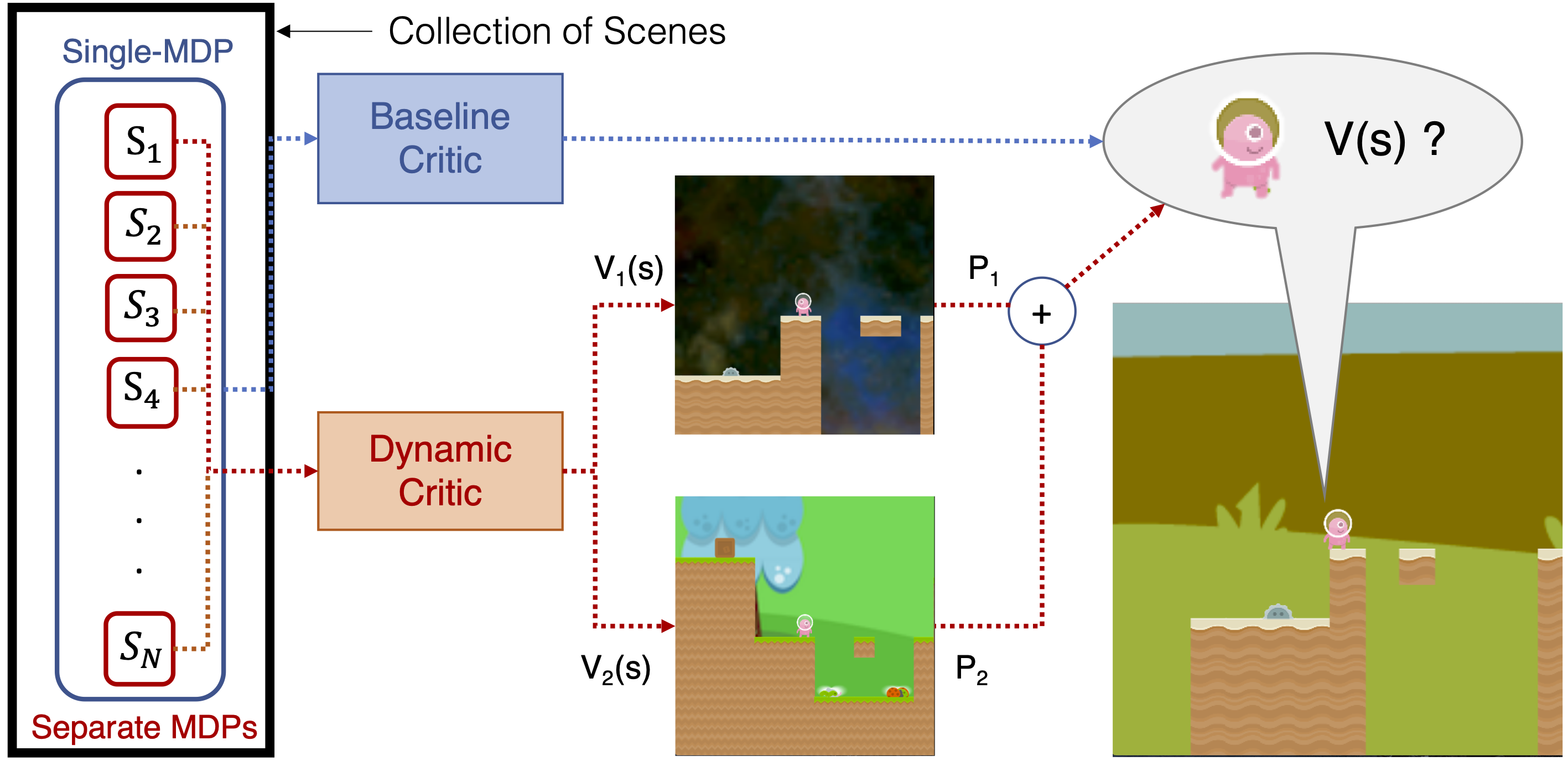}}
\caption{Traditional methods treat the set of training scenes as a single MDP and thus learn a single scene-generic value function estimate. In this paper, we argue that a more accurate scene-specific value function can be learned by explicitly treating each scene as a separate MDP. Our dynamic model, compresses the information from these scenes into value estimates for a few representative/ basis MDPs. These basis estimates are then combined based on the similarities between the current and the basis MDPs, to yield the final value function estimate. Refer section \ref{our_method} for a mathematical treatment of this concept.}
\label{fig:cartoon_intro}
\end{center}
\vskip -0.2in
\end{figure}

While the field of reinforcement learning has shown tremendous progress in the recent years, generalization across variations in the environment dynamics remains out of reach for most state-of-the-art deep RL algorithms. In order to achieve the generalization objective, many deep RL approaches attempt to train agents on environments comprising of multiple levels or scenes from the same task \cite{wortsman2019learning,cobbe2019quantifying,zhu2017target,justesen2018illuminating,cobbe2019leveraging}.  Although incorporating a wide source of data distribution in the training itself has shown promise in bridging the train and test performance, the inclusion of multiple scenes, each defined by a distinct underlying MDP, significantly increases the variance of samples collected for policy gradient computations \cite{song2019empirical}.

The current approaches using multi-level environments for training usually deal with the high variance problem by deploying multiple actors for collecting a larger and varied range of samples. For instance, \cite{zhu2017target,wortsman2019learning} use multiple asynchronous actor critic (A3C) models when training on the AI2-THOR framework  \cite{kolve2017ai2} based visual navigation task. Similarly, \cite{cobbe2019leveraging,cobbe2019quantifying} deploy parallel workers to stabilize policy gradients for multi-level training on procedurally-generated game environments. In most of these methods, a significant drop in both stability and final performance is observed on reducing the number of parallel workers.

As we will show in section \ref{variance_reduction}, the theoretical lower bound for sample variance in multi-scene RL can be achieved by learning a separate value function for each scene.  However, given the lack of information about the operational level at test times, most RL generalization benchmarks \cite{nichol2018gotta,zhang2018study,cobbe2019quantifying,igl2019generalization} \emph{effectively} treat the collection of scenes as a \emph{single-MDP environment}. That is, a common and scene generic value function is learned across all levels. We instead argue that a more scene-specific value function estimate can be learned by acknowledging each scene as a separate MDP. For the latter treatment, we refer to the corresponding collection of scenes as a \emph{multiple-MDP environment}.

In this paper, we propose a dynamic critic model, based on the clustering observed in the value function distribution across different scenes. This results in the agent being able to learn a much better MDP-specific value function (and hence the advantage function), leading to lower sample variance for policy updates. The final proposed model mimics a memory fallback strategy frequently invoked in human learning, where the learning agent falls back on its knowledge about critical but similar experiences (MDPs) to judge the current situation. Figure \ref{fig:cartoon_intro} presents an overview of our method.

The main contributions\footnote{We would like to emphasize that though multi-scene leaning helps with generalization, it is not the focus of this paper. Instead, we are interested in learning a value function that achieves the theoretical lower bound for policy gradient sample variance across different scenes (refer Eq. \ref{eq:var2}).} of this paper are as follows:
\begin{itemize}
\item Demonstrate the theoretical shortcoming of the traditional value estimation approach for \emph{multiple-MDP environments}.
\item Demonstrate that the true value function distribution forms clusters with multiple modes, that are not fully captured by the current CNN or LSTM based critic networks.
\item Propose an alternative critic model for enforcing multi-modal learning, followed by extensive analysis for enhanced variance reduction in multi-scene reinforcement learning.
\end{itemize}

\section{Preliminaries}

\subsection{Problem Setup}
A typical reinforcement learning problem consists of an environment defined by a single MDP $\mathcal{M}$ with state space $\mathcal{S}$, transition probabilities $P(s_{t+1}|s_t,a_t)$, reward function $r(s_{t},a_t,s_{t+1})$ and discount factor $\gamma$. An agent with action space $\mathcal{A}$ interacts with the environment by observing state $s_t$ at each time step $t$ and responds with an action using policy $\pi(a|s)$, resulting in a reward of $r(s_{t},a_t,s_{t+1})$. This process continues until a terminal/ absorbing state is reached and the episode ends. The goal of the RL agent is to maximize the cumulative reward $\mathcal{R}=\sum_{t=0}^{T-1} \gamma^{t} r(s_{t},a_t,s_{t+1})$.

In contrast, a multiple-MDP environment is characterized by a set of possible MDPs ${\mathcalbf{M}}: \{\mathcal{M}_1,\mathcal{M}_2, ... \mathcal{M}_N\}$, each defined by its own state space $\mathcal{S}_\mathcal{M}$, transition probabilities $P_\mathcal{M}(s_{t+1}|s_t,a_t)$ and reward function $r_\mathcal{M}(s_{t},a_t,s_{t+1})$. An episode trajectory $\tau: (s_0,a_0,s_1 ... s_T)$ can thus be sampled using any of the possible MDPs from the set $\mathcalbf{M}$, with partial or no knowledge of the agent. The goal of the agent in this case would be to maximize the expected trajectory rewards for the entire set $\mathcalbf{M}$, i.e. $\mathbf{E}_{\tau,\mathcal{M}}\left[\mathcal{R}_{\tau,\mathcal{M}}\right]$.

\subsection{Variance Reduction in Policy Gradient Algorithms}
\label{variance_reduction}

For a single-MDP environment, with policy network $\pi$ (parameterized by $\theta$) and an action-value function $Q(s,a)$, a general expression for computing policy gradients with minimal possible sample variance \cite{greensmith2004variance,schulman2015high} is given by ,
\begin{equation}
\nabla_{\theta} J = \mathbf{E}_{s,a} \left[ (\nabla_{\theta} \log\pi(a|s)) \ \psi(s,a) \right], \label{eq:policy_grad_psi}
\end{equation}

where $\psi(s,a) = Q(s,a) - V(s)$ is known as the advantage function. Similarly for multiple-MDP environments, it can be shown that the optimal formulation for minimizing sample variance is given by\footnote{We refer the readers to the supplementary materials for the mathematical proof.},
\begin{equation}
\nabla_{\theta} J = \mathbf{E}_{s,a,\mathcal{M}} \left[ (\nabla_{\theta} \log\pi(a|s)) \ \psi(s,a,\mathcal{M}) \right], \label{eq:policy_grad_psi_multi}
\end{equation}

where $\psi(s,a,\mathcal{M}) = Q(s,a,\mathcal{M}) - V(s,\mathcal{M})$. Here,  $Q(s,a,\mathcal{M}) \ \& \  V(s,\mathcal{M})$ represent the action-value and value function respectively for the particular MDP $\mathcal{M}$. However, since most of the times knowledge about the operational MDP $\mathcal{M}$, is unknown to the agent, the current policy gradient methods continue to use a single value function estimate $\hat{V}(s)$ for variance reduction, which is an estimate of the global average over the underlying value function estimates $\{V_{\mathcal{M}_1}(s),V_{\mathcal{M}_2}(s), ... , V_{\mathcal{M}_N}(s)\}$. 

We now show that such a simplification is not necessary and by learning a posterior on the set of a few representative MDPs, it is possible to learn a much better estimate for $V(s,\mathcal{M})$. 
\section{Our method}
\label{our_method}

\subsection{Motivation: Clustering Hypothesis}
\label{clustering_hypothesis}
Unlike multi-task RL, where the agent is trained on a collection of loosely related tasks, we usually want to restrict the set of MDPs to a single domain for multi-scene learning. In the case of \emph{arcade gaming}, the domain task would be a particular game, while the set of MDPs would correspond to the various levels. For the task of \emph{visual navigation}, the set of MDPs would correspond to different room conditions and setup.

\textbf{Hypothesis.} If the set of MDPs $\mathcalbf{M} : \{\mathcal{M}_1,\mathcal{M}_2, ..., \mathcal{M}_N\}$ is related through a common domain task and have similar state spaces $\mathcal{S_M}$, the value function estimates $\{V_{\mathcal{M}_1}(s),V_{\mathcal{M}_2}(s), ... , V_{\mathcal{M}_N}(s)\}$ for a given state $s$ would form clusters.

\begin{figure}[ht]
\begin{center}
\centerline{\includegraphics[width=\linewidth]{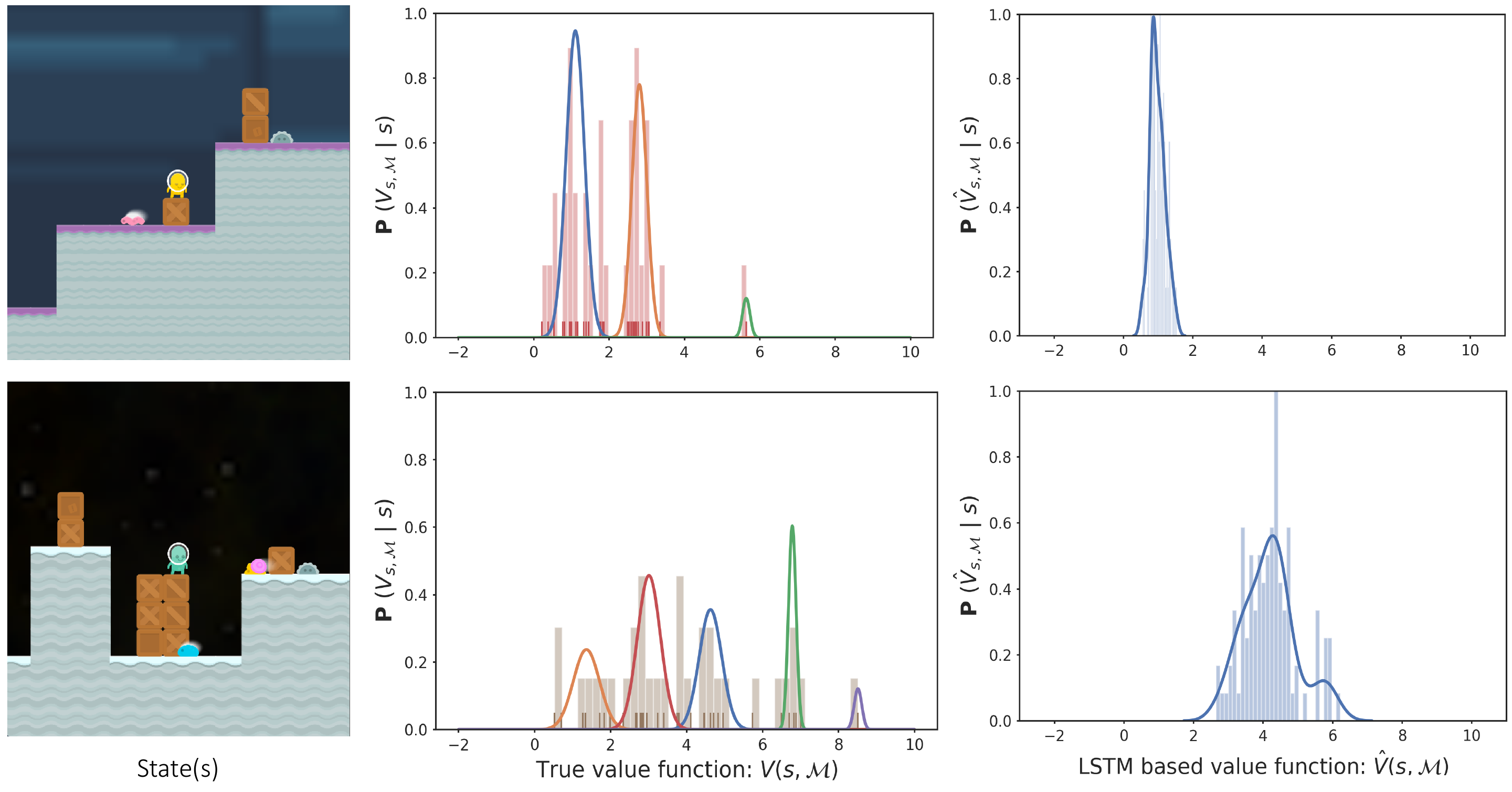}}
\caption{(Column 1-2) Results showing (true) value function clustering for an intermediate policy $\pi$, on a set of randomly selected 50 levels from the CoinRun environment. The true value estimate for an agent observing a state image on the left can be characterized by one of the many formed clusters. (Column 3) In contrast, the LSTM based value predictions, though showing some variance with MDP $\mathcal{M}$, fail to capture the multiple dominant modes exhibited by the true value function distribution.}
\label{fig:clustering_hypothesis}
\end{center}
\vskip -0.2in
\end{figure}

\textbf{Empirical Testing.} To test our hypothesis, we follow the true value function estimation strategy from Section \ref{mdp_specific_vf_est}, to obtain 50 MDP-specific value function estimates $\{V(s,\mathcal{M}_1),V(s,\mathcal{M}_2), ... ,V(s,\mathcal{M}_{50})\}$ corresponding to a random selection of 50 levels from the CoinRun ProcGen environment \cite{cobbe2019quantifying}. We then use a Gaussian Mixture Model (GMM) for fitting these $V(s,\mathcal{M}_i) \{i \in [1,50]\}$ estimates. Results are shown in Fig. \ref{fig:clustering_hypothesis}. We clearly observe that the value function estimates form multiple clusters that are not captured by traditional CNN or LSTM based critic networks.

\subsection{Dynamic Value Function Estimation}
\label{dynamic_vf_estimation}

The sample variance ($\nu$) of policy gradients defined by Eq. \ref{eq:policy_grad_psi_multi}, can be approximated as,
\begin{align}
    \nu  \approx \kappa \  . \ \mathbf{E}_{s,a,\mathcal{M}} \left[\psi^2(s,a,\mathcal{M})\right] = \kappa  \ . \  \mathbf{E}_{s,a,\mathcal{M}} \left[\left(Q(s,a,\mathcal{M})-\hat{V}(s,\mathcal{M})\right)^2\right], 
    \label{eq:var1}
\end{align}

where $\kappa = \mathbf{E}_{s,a,\mathcal{M}} \left[\left(\nabla_{\theta} \log\pi(a|s)\right)^2\right]$ and $\hat{V}(s,\mathcal{M})$ represents the predicted value function. Now, using the true value function $V(s,\mathcal{M}) = \mathbf{E}_{a}\left[Q(s,a,\mathcal{M})\right]$, Eq. \ref{eq:var1} can be decomposed as,
\begin{align}
    \nu  & \approx \kappa  \ . \  \mathbf{E}_{s,a,\mathcal{M}} \left[\left(Q(s,a,\mathcal{M}) - V(s,\mathcal{M}) \  + \  V(s,\mathcal{M}) - \hat{V}(s,\mathcal{M})\right)^2\right] \\
    & \approx \underbrace{\kappa  \ . \  \mathbf{E}_{s,a,\mathcal{M}} \left[\left(Q(s,a,\mathcal{M})-V(s,\mathcal{M})\right)^2\right]}_{\text{minimal possible variance}} + 
    \underbrace{\kappa  \ . \ \mathbf{E}_{s,\mathcal{M}} \left[\left(V(s,\mathcal{M}) - \hat{V}(s,\mathcal{M})\right)^2\right]}_{\text{prediction error}} +
    \cancelto{0}{(\dots)}. \label{eq:var2}
\end{align}

Thus, the policy gradient sample variance can be minimized by reducing the error between the true value function $V(s,\mathcal{M})$ and the predicted estimate $\hat{V}(s,\mathcal{M})$. While the exact estimation of the true value function is infeasible without knowledge of MDP $\mathcal{M}$, we use the results of clustering hypothesis in Section \ref{clustering_hypothesis}, to assert that the prediction error can be reduced by approximating the value function as the mean value of the cluster to which the current MDP belongs. 

Using the results from Fig. \ref{fig:clustering_hypothesis}, it is natural to model the true value function as a Gaussian Mixture Model (GMM) like,
\begin{equation}
P \left(V(s_t,\mathcal{M}) \vert s_t \right) = \sum_{i=1}^{N_b} p_i \ \mathcal{N}(V(s_t,\mathcal{M})|\mu_i,\sigma_i^2).
\label{eq:gmm_model}
\end{equation}

For the current state, the probability ($\alpha_i$) of  MDP $\mathcal{M}$ belonging to the $i^{th}$ cluster can be computed using the Bayes rule as,
\begin{equation}
    \alpha_i (s_t,\mathcal{M}) = \frac{p_i \  \mathcal{N}(V(s_t,\mathcal{M})|\mu_i,\sigma_i^2)}{\sum_i p_i \  \mathcal{N}(V(s_t,\mathcal{M})|\mu_i,\sigma_i^2)}.
\end{equation}

If we approximate each cluster with its mean value, we can model the predicted value estimate as,
\begin{align}
\hat{V}(s_t,\mathcal{M}) = \sum_{i=1}^{N_b} \alpha_i(s_t, \mathcal{M}) \ \mu_i(s_t, \mathcal{M}).
\label{eq:gmm_mean}
\end{align}

However, we note that $\alpha_i,\mu_i$ are parameters of the true value function distribution and not available to us during training. Nonetheless, they can be learned 
directly from the state-trajectory pairs $\{s_t,\tau^{t-}\}$,  ($\tau^{t-} : \{s_0,s_1,s_2 .... s_{t-1}\}$ is the trajectory till time $t-1$) using a Long Short Term Memory (LSTM) network. 

Interestingly, \cite{duan2016rl} had previously proposed the use of recurrent networks and episode trajectories for adapting to environment dynamics. However, as shown in Fig. \ref{fig:clustering_hypothesis}, the vanilla-LSTM based value function is usually characterized by a single dominant mode, and fails to capture the multi-modal nature of true value function distribution $V(s,\mathcal{M})$. Our method explicitly forces multiple dominant modes while estimating the cluster means $\mu_i$ and uses episode trajectories to compute assignment of the current state sample to each cluster.

\subsection{Interpretation as Learning Basis MDPs}
\label{interpretation_basis}

The final dynamic value function estimation model resulting from Eq. \ref{eq:gmm_mean} can also be interpreted as learning a set of basis value function estimates $\{\mu_1,\mu_2, ... , \mu_{N_b}\}$. These estimates can be further thought of as estimates of the value function belonging to a set of basis MDPs $\mathcalbf{M}_b:\{\mathcal{M}_{b_1},\mathcal{M}_{b_2}, ... , \mathcal{M}_{b_{N_b}}\}$, which might or might not be subset of the original MDP set $\mathcalbf{M}$. 
Eq. \ref{eq:gmm_mean} can thus be written as,
\begin{equation}
\hat{V}(s_t,\mathcal{M}) = \sum_{i=1}^{N_b} P(\mathcal{M}_{b_i} | s_t,\tau^{t-} ) \ \hat{V}(s_t,\mathcal{M}_{b_i}).
\label{eq:basis_mean}
\end{equation}

Intuitively speaking, given a state of confusion, the agent relies on the value function for the basis MDPs along with its past experience from the current episode to form an estimate of the value function for the current state. This stems from the fact that not all levels of a game are critical for learning and often have repeated situations. The set of basis MDPs thus reflects a set of levels with varied patterns that are critical for efficient game play.

\section{Evaluation on OpenAI Procgen Environments}

\subsection{Experimental design}
\begin{figure}[ht]
\begin{center}
\centerline{\includegraphics[width=\columnwidth]{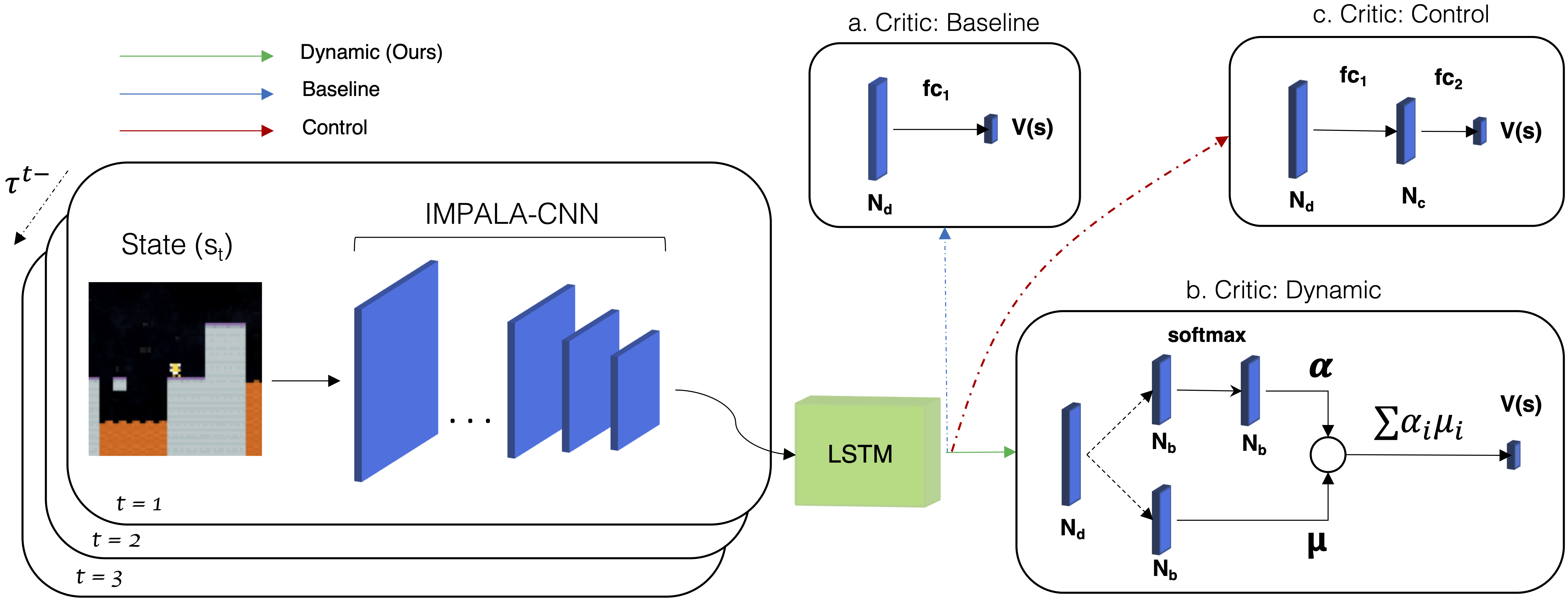}}
\caption{Illustration of model design for evaluating performance on the ProcGen environments for different configurations of the critic network:  baseline (a), dynamic (b) and control (c). The control network (c) uses a second hidden layer to compensate for the additional parameters introduced in the dynamic model (b).}
\label{fig:procgen_model}
\end{center}
\vskip -0.1in
\end{figure}

In order to do a fair estimate of the benefits of the dynamic model, we adhere to the following three configurations (Fig. \ref{fig:procgen_model}) for training and testing:

\textbf{Baseline RL$^2$.} The baseline model (Fig. \ref{fig:procgen_model}a.) closely follows\textsuperscript{\ref{fn:lstm_baseline}} the one described in \cite{cobbe2019leveraging}. Both actor and critic share an IMPALA-CNN network \cite{espeholt2018impala} modeled in the form of an LSTM \cite{duan2016rl}. The output from the LSTM is then fed to both actor and critic separately, and is followed by a single fully-connected (FC) layer to compute the action scores and value function estimation, respectively. The choice of hyper parameters is kept similar to that described in \cite{cobbe2019leveraging} in order to allow for comparison on the baseline values \footnote{\label{fn:lstm_baseline} \textit{Cobbe et al.}\cite{cobbe2019leveraging} use an IMPALA CNN network as the baseline, while we use an IMPALA CNN-LSTM network \cite{duan2016rl}.}, though, local hyper-parameter search was performed for each individual game to achieve the best average reward score.

\textbf{Dynamic.}  The dynamic model (Fig. \ref{fig:procgen_model}b.) is quite similar to the baseline and only requires few changes in the critic network to model the dynamic estimate described by Eq. \ref{eq:basis_mean}. The probabilities  $P(\mathcal{M}_{b_i} | s_t,\tau^{t-} )$ are modeled using a fully connected layer followed by a softmax function, while the mean estimates $\hat{V}(s,\mathcal{M}_{b_i})$ are learned using a single fc layer.
The choice of hyper-parameter $N_b$ is empirically determined using trial and error. For most ProcGen environments, the optimal choice for $N_b$ was found to lie within the range $[2,10]$. In section \ref{find_ncluster}, we demonstrate another way of determining the optimum choice for the number of basis clusters $N_b$.

\textbf{Control.} We design the control network (see Fig. \ref{fig:procgen_model}c.) to make sure that the performance gain from the dynamic model is not resulting from either (slight) increase in the number of parameters ($0.41 \%$ increase) or changes in hyper-parameter selection. The increased number of parameters is compensated by adding a hidden layer with size $N_c = 2N_b$ in the critic network.

All 3 configurations are trained using the Proximal Policy Optimization (PPO \cite{schulman2017proximal}) algorithm. The algorithm is ran with 4 parallel workers for gradient computations as this is seen to enhance performance. Each worker is trained for 50M steps, thus equating to a total of 200M steps across all the 4 workers. Unless otherwise specified, the final reward scores are reported as the average across 4 runs and use 500 levels for training.

\subsection{Results}
\begin{figure}[ht]
\begin{center}
\centerline{\includegraphics[width=\columnwidth]{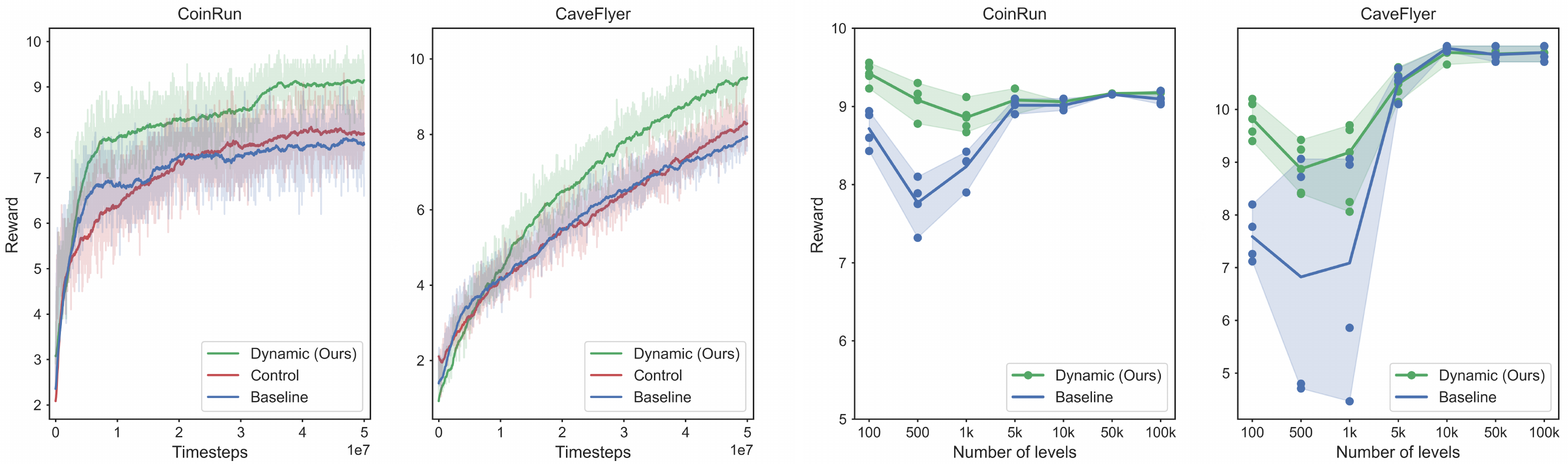}}
\caption{(Left) Performance comparison for baseline, dynamic and control configurations of the critic network, illustrating differences in sample efficiency and total reward score. (Right) Performance comparison trend over the number of training levels.}
\label{fig:procgen_train_curves}
\end{center}
\vskip -0.1in
\end{figure}

Despite the simplicity of the dynamic model, our method consistently outperforms the standard baseline models over a range of procgen environments. The improvements in performance can be seen in terms of both sample efficiency and the final reward score in (Fig. \ref{fig:procgen_train_curves}). For instance, our method results in an $18.2 \%$ and $32.3 \%$ increase in the average episode reward on the CoinRun and CaveFlyer environments, respectively. Finally, the marginal to none gains seen with the control network demonstrate that the model complexity does not trivially depend on the number of parameters. The dynamic critic network, thus, with its theoretically motivated design from Eq. \ref{eq:basis_mean}, is better able to model the underlying value function distribution $V(s,\mathcal{M})$ across multiple scenes.


We also report the trend between dynamic model gains and the number of training levels in Fig. \ref{fig:procgen_train_curves}. On average, the dynamic model leads to significant performance gains till 1k training levels. After this point, generalization effect kicks in, while the variance introduced in the value function estimates due to the increased number of levels saturates \footnote{We refer the readers to the supplementary materials for further explanations.}. For the sake of completeness, we report the final scores for the baseline and dynamic models on 10 ProcGen environments in Table \ref{tab:procgen_comparison}.

\begin{table}[t]
\caption{Final reward score comparison between PPO and dynamic PPO on 10 ProcGen environments.}
\label{tab:procgen_comparison}
\vskip 0.15in
\begin{center}
\begin{small}
\begin{sc}
\begin{tabular}{lccc}
\toprule
Environment & PPO & Dynamic PPO \\
\midrule
CoinRun  & 7.75 & \textbf{9.16} \\
CaveFlyer  & 6.82 & \textbf{9.02} \\
Plunder  & 5.88 & \textbf{7.21} \\
BigFish & 15.41 & \textbf{18.20} \\
Jumper  & \textbf{6.61} & 6.52 \\
Chaser  & 7.41 & \textbf{9.64} \\
Climber & 7.50 & \textbf{8.14} \\
DodgeBall & 9.85 & \textbf{10.14} \\
FruitBot  & 4.21 & \textbf{10.47} \\
BossFight & 10.33 & \textbf{10.75} \\
\bottomrule
\end{tabular}
\end{sc}
\end{small}
\end{center}
\vskip -0.1in
\end{table}
    
\section{Evaluation on the Visual Navigation Task}
\subsection{Task Definition}
The task of visual navigation consists of a set of scenes $\mathcalbf{S} = \{\mathcal{S}_1, \mathcal{S}_2, ... \mathcal{S}_n\}$, and a set of possible object classes $\mathcalbf{O} = \{\mathcal{O}_1,\mathcal{O}_2 .. \mathcal{O}_m\}$. Note that the set $\mathcalbf{S}$ is just another annotation for the MDP set $\mathcalbf{M} = \{\mathcal{M}_1, \mathcal{M}_2, ... \mathcal{M}_n,\}$ as each scene is characterized by a distinct underlying Markov Decision Process (MDP). Each scene $\mathcal{S}_i$ can have different room setup, distribution of objects and lighting conditions.

A single navigation task $\mathcal{T}$ consists of an agent with action space $\mathcal{A}$ situated in a random position $p$ within the one of the scenes $\mathcal{S}_i$. The goal of the agent is to reach an instance of the target class $\mathcal{O}_k$ (given as a Glove embedding \cite{pennington2014glove}) within a certain number of steps. The agent then continues interacting with the environment using a policy $\pi_\theta$ until it chooses a termination action. The episode is considered a success, if and only if, at the time of termination, the target object is sufficiently close and in the field of view of the agent.

\subsection{Experimental design}
We use the following 3 metrics for evaluating the results on this task.

\textbf{SPL.} Success weighted by Path Length (SPL) proposed by \cite{anderson2018evaluation} measures the navigation efficiency of the agent and is given by $\frac{1}{N} \sum_{i=1}^N S_i \frac{L_i}{\max(P_i,L_i)}$.

\textbf{Success rate.} The average rate of success: $\frac{1}{N} \sum_{i=1}^N S_i$.

\textbf{Total reward.} Average total episode reward: $\frac{1}{N} \sum_{i=1}^N \mathcal{R}_i$.

In above, $N$ is the number of episodes, $S_i \in \{0,1\}$ indicates the success of an episode, $P_i$ is the path length, $L_i$ is the optimal path length to any instance from the target object class in that scene, and $R_i$ is the episode reward. 
The baseline results are reported using the non-adaptive and self-adapting (SAVN) A3C models from \cite{wortsman2019learning} with 12 asynchronous workers. We then modify the critic network  as per Eq. \ref{eq:basis_mean} to get the dynamic version for both baselines. 


The agent is trained using the  AI2-THOR environment \cite{kolve2017ai2} which consists of 120 distinct scenes. A train/val/test split of 80:20:20 is used for selecting the best model based on success rate. Note that the final results are reported after 5M episodes of training which equates to $\approx$ 72 hours of training time on 2 GeForce RTX 2080 Ti GPUs.

\subsection{Results}

\begin{table}[ht]
\label{tab:visnav}
\begin{center}
\begin{small}
\begin{tabular}{lcccl}
\toprule
Method & SPL & Success & Total Reward \\
\midrule 
A3C  & 14.3 & 31.8 & 1.413 \\
Dynamic A3C & \textbf{15.4} &  \textbf{36.5} & \textbf{1.638}\\
\midrule 
SAVN & \textbf{15.19} & 37.1 & 1.652 \\
Dynamic SAVN & 14.81 &  \textbf{38.7} & \textbf{1.824}\\
\bottomrule
\end{tabular}
\end{small}
\end{center}
\caption{Comparison on key evaluation metrics for visual navigation. We observe improvements when using the dynamic critic network on both the non-adaptive and adaptive baselines.}
\vskip -0.1in
\end{table}

\begin{figure}[t]
\vskip 0.2in
\begin{center}
\centerline{\includegraphics[width=0.5\columnwidth]{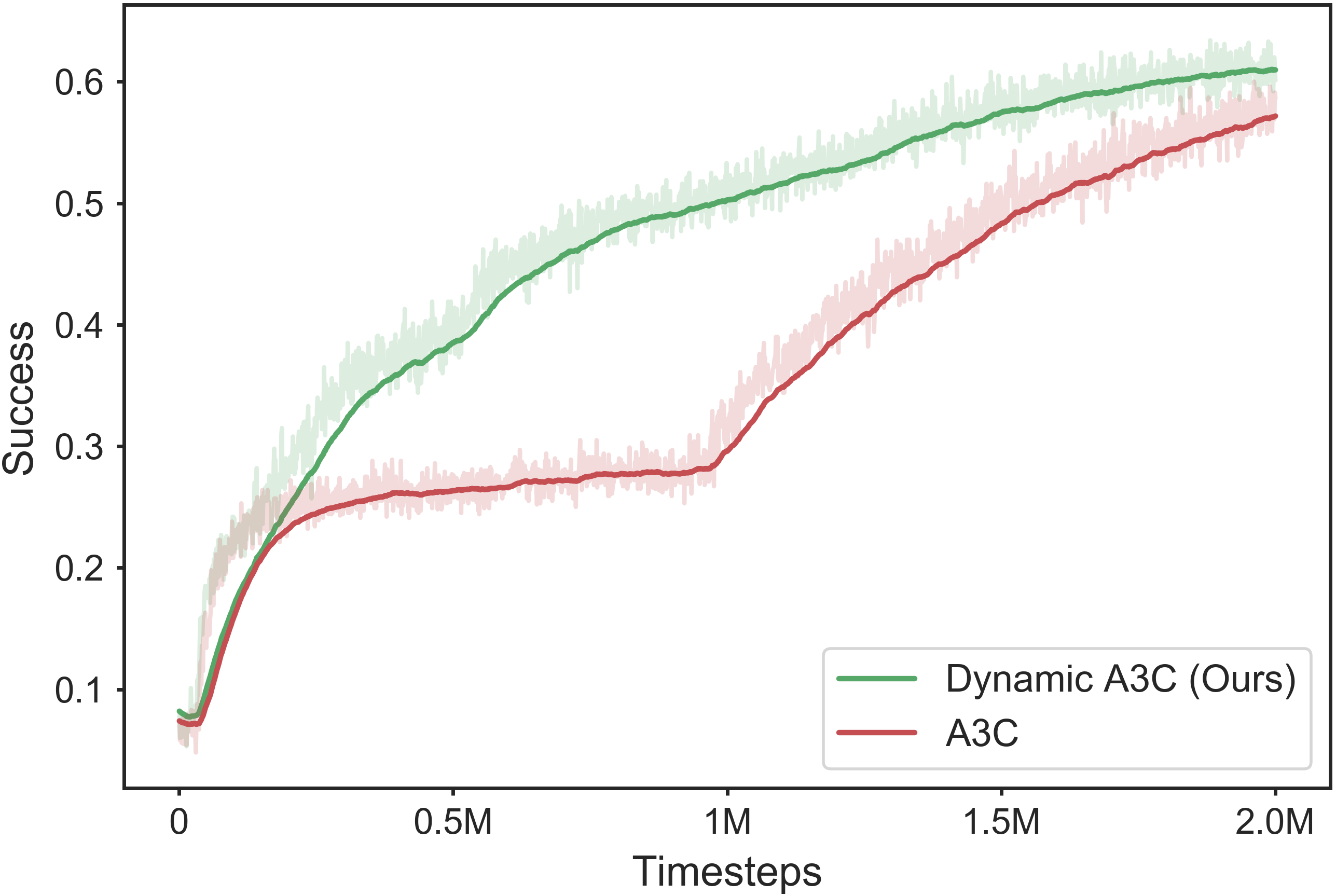}}
\caption{Comparing sample efficiency at training time for A3C and Dynamic A3C model on the visual navigation task.}
\label{fig:visnav_train_curve}
\end{center}
\vskip -0.2in
\end{figure}

As described in Table \ref{tab:visnav}, the dynamic A3C model results in significant improvements across all 3 performance metrics for both self-adaptive and non-adaptive baselines. Furthermore, the dynamic model provides a huge boost in the sample efficiency (Fig. \ref{fig:visnav_train_curve}).
The dynamic nonadaptive A3C model reaches a test success rate of $\approx 30$\% after only 2M episodes which in contrast with the non-adaptive baseline takes around 4M episodes. Similarly, the dynamic model for adaptive A3C achieves a test success rate of $\approx 40$\% after only 1M episodes while the corresponding baseline has a success rate of only $30.8$\%. Given the huge amounts of training time required in simulation, the dynamic model can be a real asset for applications with strict time constraints. For instance, the high sample efficiency is useful for quick fine-tuning in mobile navigation robots operating in the real world \cite{chancan2019visual}.

\section{Analysis}
\subsection{Method: Estimating True MDP-Specific Value Function}
\label{mdp_specific_vf_est}

In order to make any comments about the distribution of the value function $V(s,\mathcal{M})$ over $s \in \mathcal{S} \ \& \  \mathcal{M} \in \mathcalbf{M}$, we first need to be able to compute the true scene-specific value estimates for any state $s \in \mathcal{S}$ and a particular MDP $\mathcal{M}$. While training an expert model for each level \cite{parisotto2015actor} is possible, it suffers from the following flaws:

\begin{enumerate}
\item The resulting value estimates $\{V_1(s),V_2(s), ... V_n(s)\}$ would not be over the same policy $\pi$.
\item The learned CNN weights will only be able to correctly process state images $s$ belonging to the MDP specific state space subset $\mathcal{S}_{\mathcal{M}_i}$
\end{enumerate}

We instead propose the following strategy for estimation of true MDP-specific value function:
\begin{itemize}
\item Train a general actor critic model over the entire set $\mathcalbf{M}$ to learn an intermediate but sufficiently good policy $\pi$. 
\item Freeze all the weights except the non-shared critic branch (usually a single fc layer)
\item Fine-tune this semi-frozen network, individually on each MDP $\mathcal{M}_i$ using only the critic value loss.
\end{itemize}

Thus, each of the final fine-tuned model learns a value function for  the same policy $\pi$, but thanks to the unaltered CNN weights, is also able to process state images belonging to other MDPs.

\subsection{Finding the Optimal Number of Dynamic Nodes}
\label{find_ncluster}
A critical component in training the dynamic model is the selection of number of base nodes / clusters $N_b$. While it is possible to treat it as another hyper-parameter, we present an alternate approach for the same. This approach not only gives us the number of optimal clusters, but also strengthens our belief in the function of the proposed critic network.

Recall that, given our motivation from the clustering hypothesis (Section \ref{clustering_hypothesis}), we believe that the dynamic critic network learns to model the value function distribution using a Gaussian Model Mixture (GMM). The final model needs to learn this distribution not just across different MDPs belonging to the set $\mathcalbf{M}$ but also for all states $s \in \mathcal{S}$. Thus, to find out the optimal number of gaussian clusters, we begin by approximating the multi-variate distribution $V(s,\mathcal{M})$ using discrete samples $\{s_j,\mathcal{M}_i,V(s_j,\mathcal{M}_i)\}$ for $s_j \in S \ \& \ \mathcal{M}_i \in \mathcalbf{M}$.

For the sample collection, we first obtain an intermediate policy $\pi$ and MDP specific value estimate networks $\hat{V}_i(s)$ for a random selection of 500 levels from the CoinRun ProcGen environment, using the strategy from Section \ref{mdp_specific_vf_est}. While incorporating the entire state space is infeasible, we try to minimize the error by sampling a large collection of 1000 states from the different levels using the common policy $\pi$. Next for each of these states $s_j, \ j \in [1,1000]$, we obtain the corresponding level specific value function estimates using $\hat{V}_i(s_j), i \in [1,500]$. 

\begin{figure}[t]
\begin{center}
\centerline{\includegraphics[width=0.5\columnwidth]{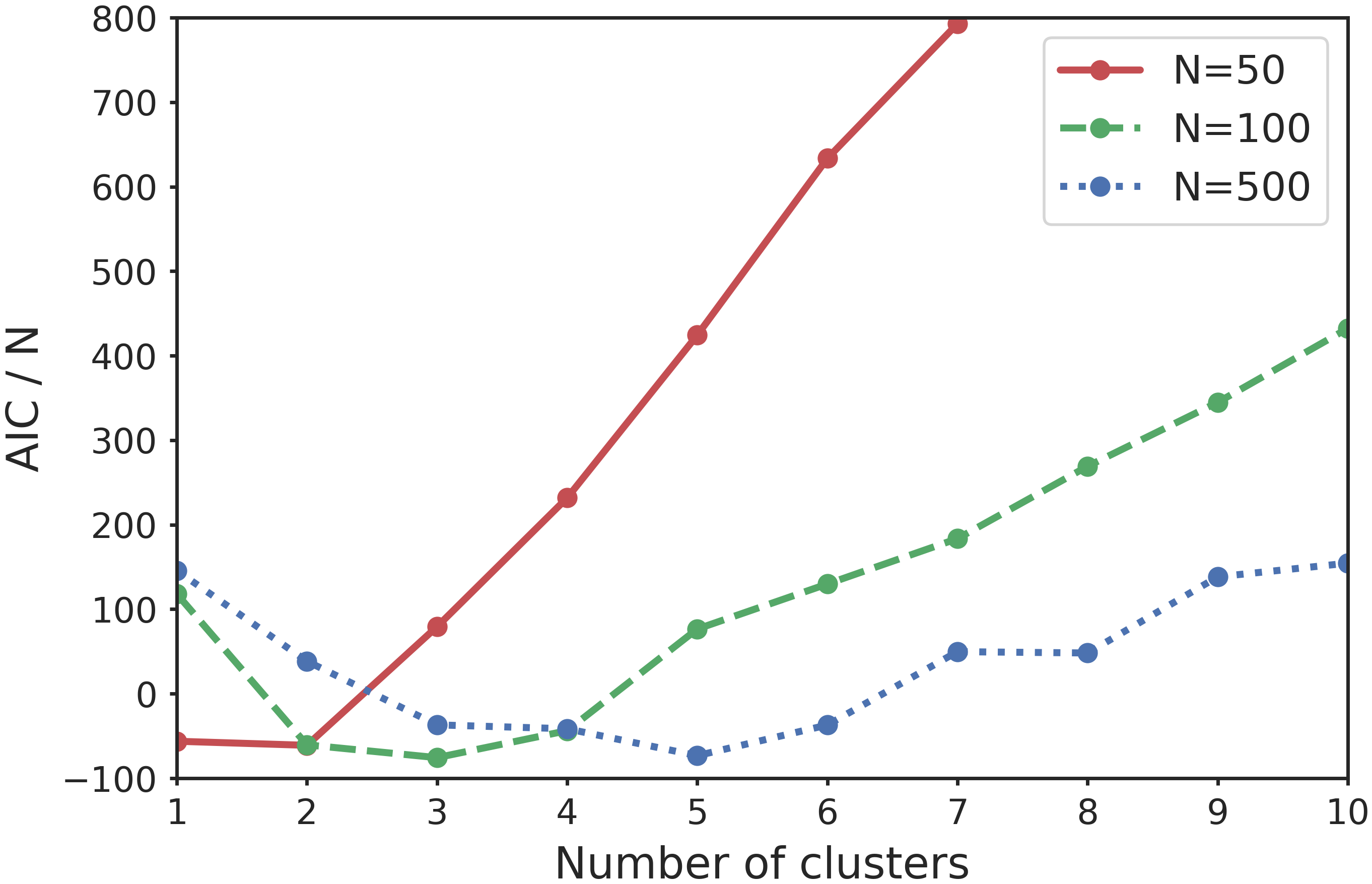}}
\caption{AIC scores for different number of clusters and training levels N. The local minima represents the Bayesian optimal choice \cite{forster2011aic} for number of basis / cluster nodes $N_b$.}
\label{fig:ncluster_aic_coinrun}
\end{center}
\vskip -0.1in
\end{figure}

The generated dataset representing samples from the multi-variate distribution $V(s,\mathcal{M})$ has shape [N x K], where N is number of levels and K is the number of states. The dataset is then fitted using a GMM with variable number of components $C \in [1,10]$. We finally use the Akaike Information Criterion (AIC) from \cite{akaike1987factor}, to determine the best selection for the number of components (Fig. \ref{fig:ncluster_aic_coinrun}). 

As shown in Fig. \ref{fig:ncluster_aic_coinrun}, the optimum number of clusters (point of minima in the AIC/N curve) increases with the number of training levels. We also note that the AIC/N curve becomes less steep, as the number of training levels increases. This implies that given a sufficient number of training levels, the dynamic model's performance shows low sensitivity (higher robustness) to the selection of the hyper-parameter $N_b$. 

\subsection{Visualizing Representative MDPs}

\begin{figure}[ht]
\begin{center}
\centerline{\includegraphics[width=\columnwidth]{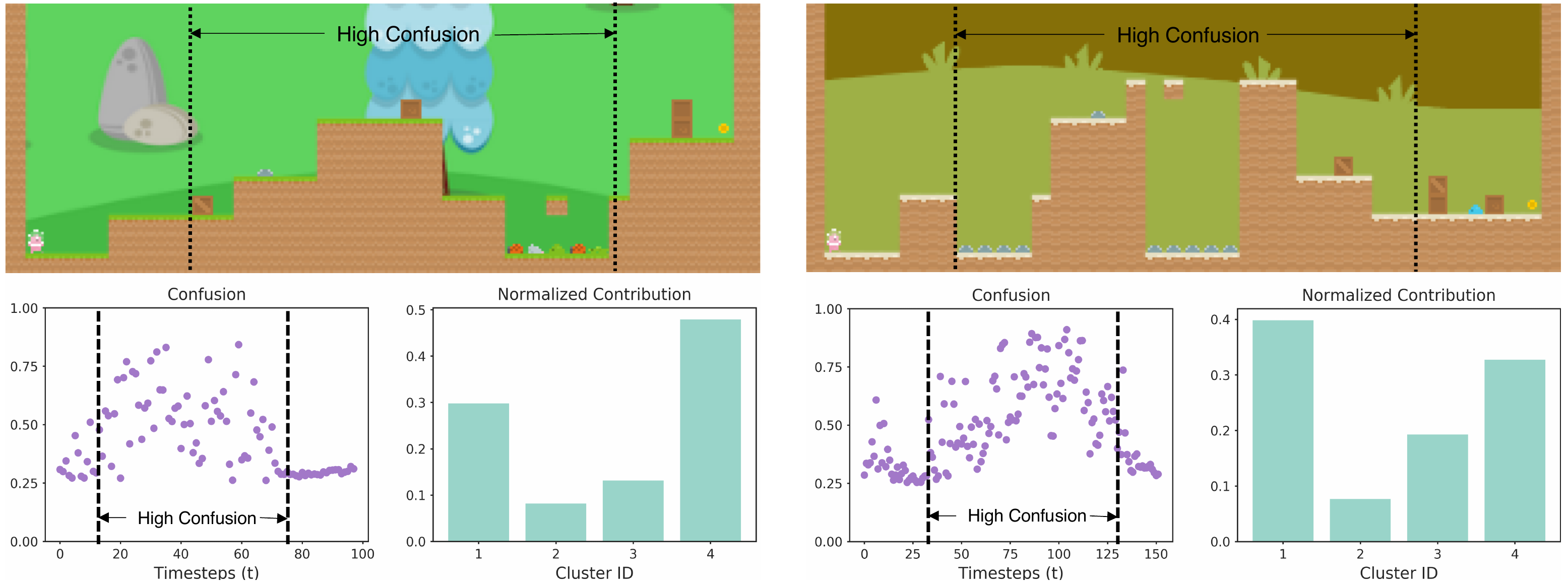}}
\caption{Analysing ``confusion'' distribution and cluster specific ``contribution'' scores for different levels from the CoinRun environment. The states with high confusion usually occur in the middle of a level where most of the obstacles are present. We also note that different clusters contribute differently to value estimation depending on the underlying scene. The cluster with highest contribution can be thought of as representing a basis MDP most similar to the current level.}
\label{fig:mdpvis}
\end{center}
\vskip -0.1in
\end{figure}

Recall from section \ref{interpretation_basis}, the dynamic model can be interpreted as learning the value function estimates for a set of basis MDPs $\mathcalbf{M}_b$, which may or may not belong to the original MDP set $\mathcalbf{M}$. In this section, we present a visualization analysis for what each basis/cluster might represent.

We first begin by defining the term ``confusion''. An agent is said to be in a state of confusion if it is unsure of the cluster/ basis MDP to which the current $\{s_t,\tau^{t-}\}$ pair belongs. As per Eq. \ref{eq:gmm_mean}, the probability of an agent (with state pair $\{s_t,\tau^{t-}\}$) choosing a particular cluster $i$ is given by $\alpha_i(s_t,\tau^{t-}), i \in [1,N_b]$. Thus, a state of complete confusion would be characterized by a uniform distribution across the the parameters $\alpha_i$, while a sharp spike in $\alpha_i$ distribution indicates low confusion. Mathematically, confusion $\delta \in [0,1]$ for the $\{s_t,\tau^{t-}\}$ pair can be defined as,
\begin{equation}
    \delta(s_t,\tau^{t-}) = \frac{1}{N_b . \sum_i \alpha^2_i(s_t,\tau^{t-})}.
\end{equation}

\begin{figure}[ht]
\floatbox[{\capbeside\thisfloatsetup{capbesideposition={right,center},capbesidewidth=7cm}}]{figure}[\FBwidth]
{\caption{High and low confusion states for different scenes from the CoinRun environment. The red and orange arrows indicate the possible paths that the agent might take to avoid the obstacle/enemy. Each path corresponds to a distinct value estimate for a basis MDP and thus, more possible paths indicate higher confusion.}
\label{fig:max_min_confusion}}
{\includegraphics[width=5cm]{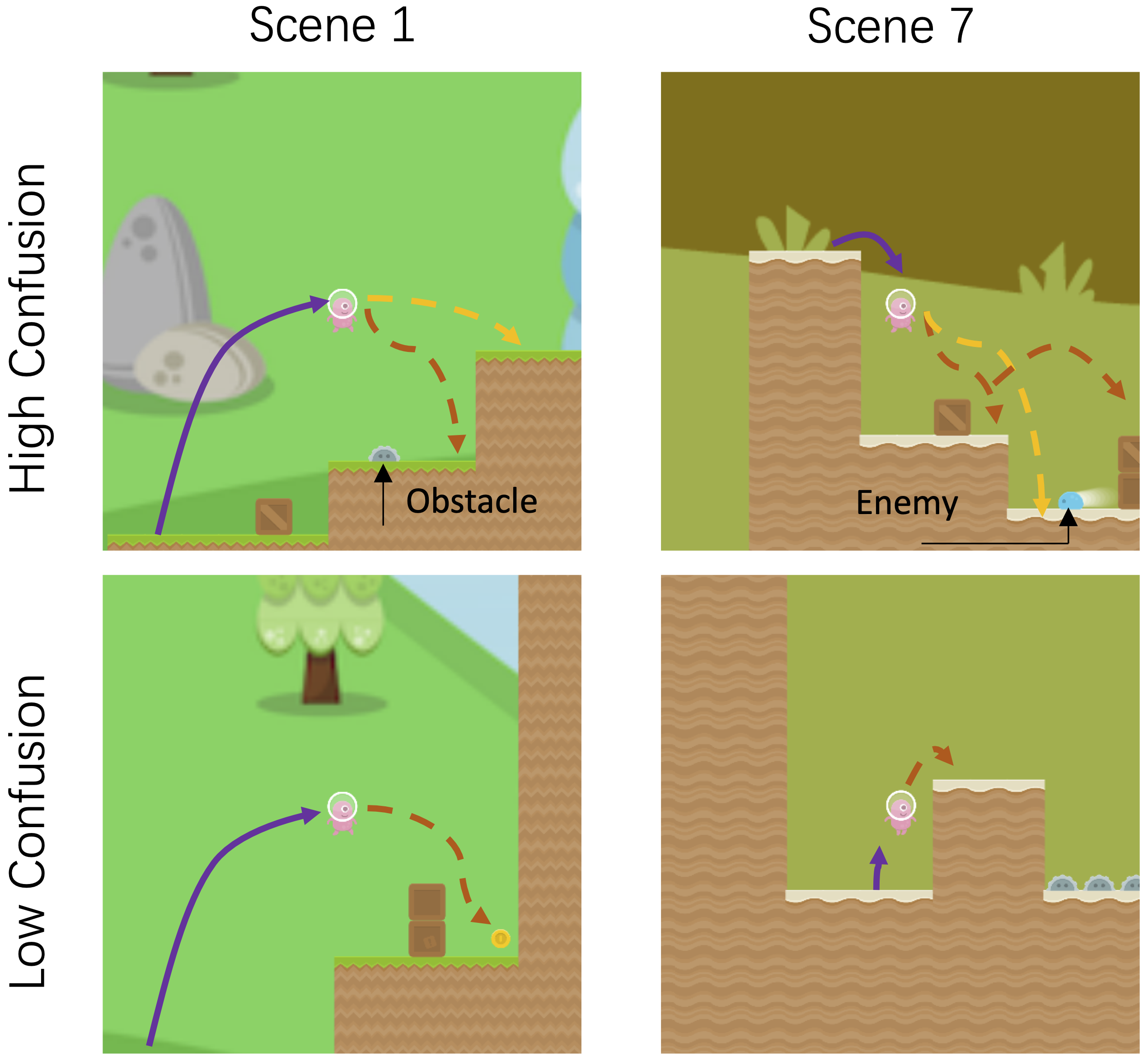}}
\end{figure}

Fig. \ref{fig:max_min_confusion} shows some states of maximum and minimum confusion for two randomly selected levels from the CoinRun environment. States with high confusion usually occur around tricky obstacles or jumps, while low confusion states usually occur in simple scenarios, happening mostly at the beginning or end of a level.

Next we define the ``contribution'' of a cluster to the value function estimation while navigating a particular scene/ level. Intuitively, a cluster with a higher average $\alpha_i$ across the episode, should have a higher contribution. We would also like to provide more weight-age to cluster contributions for states with high confusion, which as shown in Fig. \ref{fig:max_min_confusion}, are usually critical for optimal game play. Hence, given an episode trajectory $\tau: \{s_0,a_0,...,s_T\}$, we define the weighted contribution $\rho_i$ for the $i^{th}$ cluster as,
\begin{equation}
    \rho_i(\tau) = \frac{1}{T} \sum_{t=1}^{T} \delta(s_t,\tau^{t-}) \  \alpha_i(s_t,\tau^{t-}).
\end{equation}

Fig. \ref{fig:mdpvis} shows that, different clusters contribute differently to value estimation on each scene, depending on the type of obstacles present. We can further argue that the cluster with the highest contribution represents an underlying basis MDP with scene features most similar to the associated level.

\section{Conclusion}
This paper proposes a dynamic value estimation strategy for multi-scene reinforcement learning. We demonstrate that despite the lack of knowledge about the operational scene / level, a more accurate scene-specific value estimate can be learned by utilizing the clustering observed in the value function distribution across different scenes. Our dynamic critic network outperforms the current baselines in both sample efficiency and final reward across a range of multi-level ProcGen environments and the visual navigation task. Finally, we provide a mechanism for analysing the proposed model in terms of ``confusion'' and ``contribution''. This helps us visualize the representative cluster MDPs and gives deeper insights into what the dynamic agent truly learns.

 
\newpage
\bibliography{dynamic_est}
\bibliographystyle{ieeetr}

\appendix
\section{Policy Gradient Formulation for Multiple-MDP Environments}
\label{policy_grad}
The transition from single to multiple-MDP policy gradient formulation is intuitively straightforward, however, for the sake of completeness, we provide here a mathematical derivation to justify the claims made in the original paper.

\subsection{Policy Gradient Equation}
Policy gradient algorithms aim to optimize the policy $\pi(a|s)$ by maximizing the expected trajectory reward. For a multiple-MDP environment, the expected reward is over different trajectories $\tau$ and the set of possible MDPs $\mathcalbf{D}$. Thus,
\begin{equation}
    J = \mathbf{E}_{\tau,\mathcal{M}}[\mathcal{R}_{\tau,\mathcal{M}}],
\end{equation}
where $\mathcal{R}_{\tau,\mathcal{M}}$ is the total reward for a trajectory $\tau$ sampled using MDP $\mathcal{M}$. The above equation can be broken down as,
\begin{align}
    &J = \sum_{\mathcal{M}} \sum_\tau P(\tau,\mathcal{M}) \ \mathcal{R}_{\tau,\mathcal{M}}. \\
    &J  = \sum_{\mathcal{M}} P(\mathcal{M}) \left(\sum_\tau P(\tau \vert \mathcal{M}) \ \mathcal{R}_{\tau,\mathcal{M}}\right).
\end{align}

The term $\sum_\tau P(\tau \vert \mathcal{M}) \ \mathcal{R}_{\tau,\mathcal{M}}$ can be understood as the expected reward over trajectories given a fixed underlying MDP $\mathcal{M}$. We denote this conditional expectation using $J_\mathcal{M}$. Thus,
\begin{align}
    J_\mathcal{M} = \sum_\tau P(\tau \vert \mathcal{M}) \ \mathcal{R}_{\tau,\mathcal{M}} \label{eq:single_conditonal}\\
    \implies J = \sum_{\mathcal{M}} P(\mathcal{M}) \ J_\mathcal{M}.
\end{align}

Computing gradients for both sides with respect to policy parameter $\theta$:
\begin{align}
    \nabla_\theta J =  \sum_{\mathcal{M}} P(\mathcal{M}) \ \nabla_\theta(J_\mathcal{M})\label{eq:policy_grad_div}.
\end{align}

Adopting single-MDP policy gradient derivation procedure from \cite{sutton2018reinforcement} for Eq. \ref{eq:single_conditonal}, the gradient with respect to $J_\mathcal{M}$ can be written as:
\begin{align}
    &\nabla_\theta(J_\mathcal{M}) = \sum_s \sum_a P(s,a|\mathcal{M})  \ [\nabla_\theta \log \pi(a|s)] \  Q(s,a,\mathcal{M}),\label{eq:single_policy_grad_extension}
\end{align}

where $\pi(a|s)$ represents the scene-generic policy function and $Q(s,a,\mathcal{M})$ is the action value function for the specific MDP $\mathcal{M}$.
Combining Eq. \ref{eq:policy_grad_div} and \ref{eq:single_policy_grad_extension},
\begin{align}
    &\nabla_\theta J = \sum_{\mathcal{M}} P(\mathcal{M}) \ \sum_s \sum_a P(s,a|\mathcal{M})  \ [\nabla_\theta \log \pi(a|s)] \  Q(s,a,\mathcal{M}). \\
    &\nabla_\theta J = \sum_{\mathcal{M}} \sum_s \sum_a P(s,a,\mathcal{M})  \ [\nabla_\theta \log \pi(a|s)] \  Q(s,a,\mathcal{M}). \label{eq:policy_grad_prob_sum}
\end{align}

Writing Eq. \ref{eq:policy_grad_prob_sum} in the form of an expectation mean:
\begin{equation}
     \boxed{\nabla_\theta J = \mathbf{E}_{s,a,\mathcal{M}} \left[\left(\nabla_\theta \log \pi(a|s)\right) \  Q(s,a,\mathcal{M})\right]} \label{eq:policy_grad_mean}.
\end{equation}

\subsection{Variance Reduction}
The goal of variance reduction is to find a distribution with same mean as Eq. \ref{eq:policy_grad_mean}, but reduced variance so as to be able to compute the sample mean with minimum error using a limited number of samples. In this section, we derive the optimal formulation for variance reduction in \emph{multiple-MDP} environments.

\textbf{Lemma 1.} $Q(s,a,\mathcal{M})$ can be replaced by $\psi(s,a,\mathcal{M}) = Q(s,a,\mathcal{M}) - f(s,\mathcal{M})$, where $f$ is any general function, without affecting the overall policy gradient from Eq. \ref{eq:policy_grad_mean}.

\textbf{Proof:}
Let the new policy gradient be $\nabla_\theta J'$,
\begin{align*}
     &\nabla_\theta J' = \mathbf{E}_{s,a,\mathcal{M}} \left[\left(\nabla_\theta \log \pi(a|s)\right) \  \psi(s,a,\mathcal{M})\right]. \\
     &\nabla_\theta J' = \mathbf{E}_{s,a,\mathcal{M}} \left[\left(\nabla_\theta \log \pi(a|s)\right) \  (Q(s,a,\mathcal{M})-f(s,\mathcal{M}))\right]. \\
     &\nabla_\theta J' = \nabla_\theta J - \mathbf{E}_{s,a,\mathcal{M}} \left[\left(\nabla_\theta \log \pi(a|s)\right) \  f(s,\mathcal{M})\right].\\
     &\nabla_\theta J' = \nabla_\theta J - \sum_\mathcal{M}\sum_s\sum_a P(s,a,\mathcal{M}) \left(\nabla_\theta \log \pi(a|s)\right) \  f(s,\mathcal{M}).\\
     &\nabla_\theta J' = \nabla_\theta J - \sum_\mathcal{M}\sum_s P(s,\mathcal{M}) \ f(s,\mathcal{M}) \ \sum_a P(a \vert s,\mathcal{M}) \ [\nabla_\theta \log \pi(a|s)].
\end{align*}

Since we learn a common policy $\pi(a|s)$ for the entire set of MDPs $\mathcalbf{D}$, $P(a \vert s,\mathcal{M}) = \pi(a|s)$,
\begin{align*}
     &\nabla_\theta J' = \nabla_\theta J - \sum_\mathcal{M}\sum_s P(s,\mathcal{M}) \ f(s,\mathcal{M}) \ \sum_a \pi(a|s) \ \nabla_\theta \log \pi(a|s).\\
     &\nabla_\theta J' = \nabla_\theta J - \sum_\mathcal{M}\sum_s P(s,\mathcal{M}) \ f(s,\mathcal{M}) \ \left(\sum_a \ \nabla_\theta \ \pi(a|s)\right).\\
     &\nabla_\theta J' = \nabla_\theta J - \sum_\mathcal{M}\sum_s P(s,\mathcal{M}) \ f(s,\mathcal{M}) \ \left( \nabla_\theta \sum_a \ \pi(a|s)\right).\\
     &\nabla_\theta J' = \nabla_\theta J - \sum_\mathcal{M}\sum_s P(s,\mathcal{M}) \ f(s,\mathcal{M}) \ \cancelto{0}{\left( \nabla_\theta .1\right)}.\\
     &\implies \boxed{\nabla_\theta J' = \nabla_\theta J}.
\end{align*}

\textbf{Lemma 2.} The optimal function $f(s,\mathcal{M})$ which minimizes the sample variance is equal to the MDP-specific value function $V(s,\mathcal{M})$. 

\textbf{Proof:}

For a given tuple $ \{s,a,\mathcal{M}\}$, the sample value $\mathcal{X}$ is given by $\left(\nabla_\theta \log \pi(a|s)\right) \  \psi(s,a,\mathcal{M})$. Thus, the sample variance can be written down as:

\begin{align*}
     &Var(\mathcal{X}) = E[\mathcal{X}^2] - (E[\mathcal{X}])^2.\\ \\
     &\text{Given that the function $f(s,\mathcal{M})$ is parameterized by $\phi$:}\\
     &\argmin_\phi Var(\mathcal{X}) = \argmin_\phi \left(E[\mathcal{X}^2] - (E[\mathcal{X}])^2\right).\\ \\
     &\text{From \textbf{Lemma 1}, we know that the expected value of the samples remains unchanged for any function $f$.}\\
     &\implies \argmin_\phi Var(\mathcal{X}) = \argmin_\phi E[\mathcal{X}^2],\\
     &\argmin_\phi Var(\mathcal{X}) = \argmin_\phi \mathbf{E}_{s,a,\mathcal{M}} \left[\left(\left(\nabla_\theta \log \pi(a|s)\right) \  \psi(s,a,\mathcal{M})\right)^2\right].\\ \\
     &\text{Assuming independence between policy network gradients and the action value function:}\\
     &\argmin_\phi Var(\mathcal{X}) = \argmin_\phi \mathbf{E}_{s,a,\mathcal{M}} \left[\left(\nabla_\theta \log \pi(a|s)\right)^2\right] \ \mathbf{E}_{s,a,\mathcal{M}} \left[\psi^2(s,a,\mathcal{M})\right],\\
     &\argmin_\phi Var(\mathcal{X}) = \argmin_\phi \mathbf{E}_{s,a,\mathcal{M}} \left[\psi^2(s,a,\mathcal{M})\right],\\
     &\argmin_\phi Var(\mathcal{X}) = \argmin_\phi \mathbf{E}_{s,a,\mathcal{M}} \left[\left(Q(s,a,\mathcal{M})-f_\phi(s,\mathcal{M})\right)^2\right].\\
    \\ \\
     &\text{For optimal $\phi, \quad \nabla_\phi Var(\mathcal{X}) = 0$} \\
     & \implies \nabla_\phi Var(\mathcal{X}) \equiv  \nabla_\phi \ \mathbf{E}_{s,a,\mathcal{M}} \left[\left(Q(s,a,\mathcal{M})-f_\phi(s,\mathcal{M})\right)^2\right] = 0.\\
     & 2 \ \nabla_\phi f \ . \  \mathbf{E}_{s,a,\mathcal{M}} \left[f_\phi(s,\mathcal{M}) - Q(s,a,\mathcal{M})\right] = 0.\\
     &\sum_\mathcal{M} \sum_s \sum_a P(s,a,\mathcal{M}) \left[f_\phi(s,\mathcal{M}) - Q(s,a,\mathcal{M})\right] = 0. \\
     &\sum_\mathcal{M} \sum_s P(s,\mathcal{M}) \sum_a P(a \vert s,\mathcal{M}) \left[f_\phi(s,\mathcal{M}) - Q(s,a,\mathcal{M})\right] = 0. \\
     &\sum_\mathcal{M} \sum_s P(s,\mathcal{M})  \left[f_\phi(s,\mathcal{M}) - \sum_a \pi(a \vert s)\  Q(s,a,\mathcal{M})\right] = 0. \\
     &\sum_\mathcal{M} \sum_s P(s,\mathcal{M})  \left[f_\phi(s,\mathcal{M}) -  V(s,\mathcal{M})\right] = 0. \\
     &\mathbf{E}_{s,\mathcal{M}} \left[f_\phi(s,\mathcal{M}) -  V(s,\mathcal{M})\right] = 0.
    \\ \\
     &\text{The above relation has to hold for any batch sample across $s,\mathcal{M}$, which implies that:}\\
     &\boxed{f_\phi(s,\mathcal{M}) = V(s,\mathcal{M})}.
\end{align*}

\vskip 0.2in
Thus, from \textbf{Lemma 1 \& 2}, the optimal formulation leading to minimum variance for a \emph{multiple-MDP} environment is given by:
\begin{equation}
    \boxed{\nabla_\theta J = \mathbf{E}_{s,a,\mathcal{M}} \left[\left(\nabla_\theta \log \pi(a|s)\right) \  (Q(s,a,\mathcal{M})-V(s,\mathcal{M}))\right]}. \label{eq:multi_policy_grad_mean}
\end{equation}

\section{Variance Saturation with Increasing Number of Scenes}

\begin{figure}[h]
\vskip 0.2in
\begin{center}
\centerline{\includegraphics[width=\textwidth]{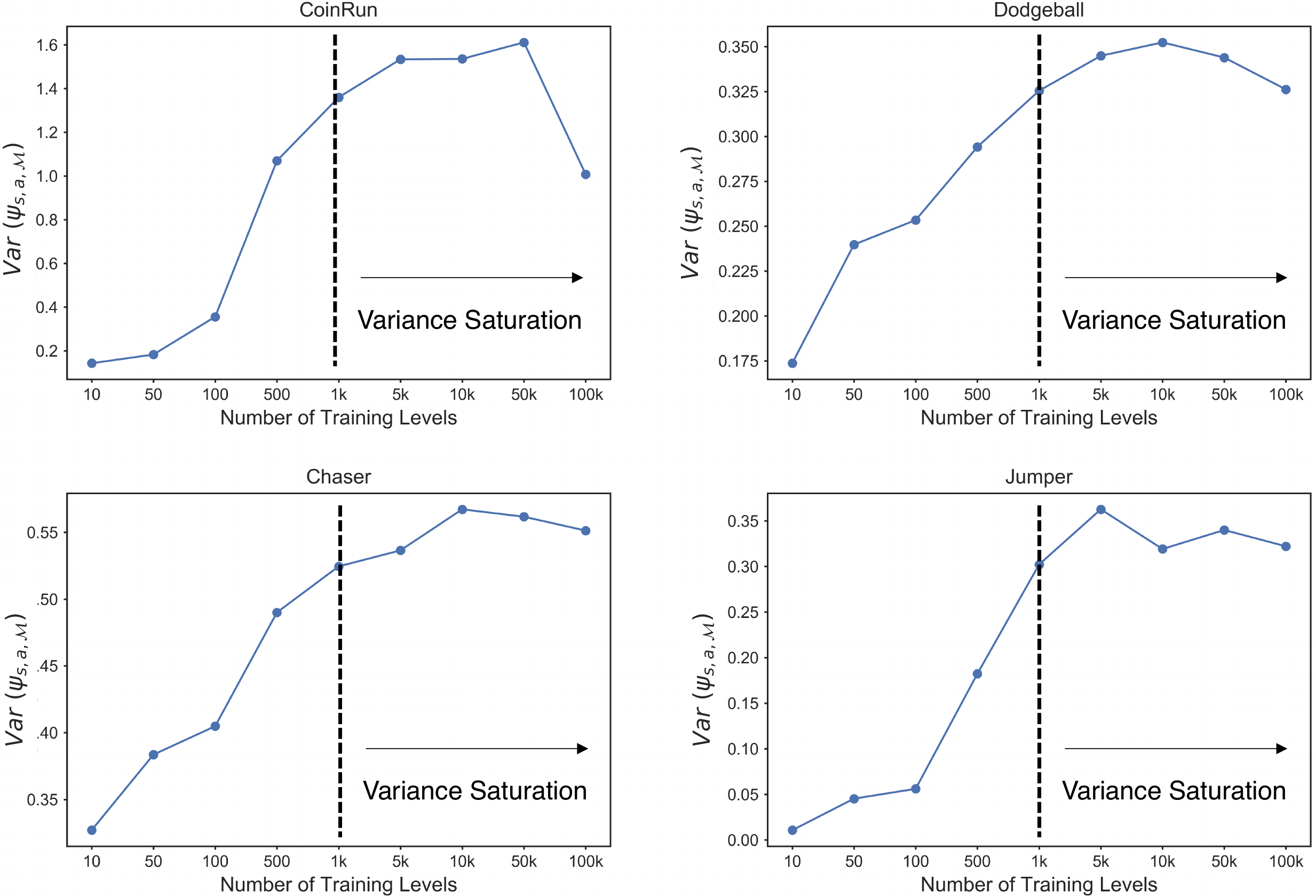}}
\caption{Demonstrating sample variance $\left(\mathbf{E}_{s,a,\mathcal{M}}[\psi^2(s,a,\mathcal{M})]\right)$ saturation with increasing number of training levels. The relative novelty of new experiences decreases as we add more scenes to the training data.}
\label{fig:variance_saturation}
\end{center}
\vskip -0.2in
\end{figure}

To determine the relationship between sample variance and the number of training levels/scenes, we analyse the equivalent sample variance $\mathbf{E}_{s,a,\mathcal{M}}[\psi^2(s,a,\mathcal{M})]$ while training the RL agent using different training level counts. We report the average sample variance across first the 5M steps on the baseline PPO (\cite{schulman2017proximal}) model in Fig. \ref{fig:variance_saturation}.

The variance introduced due to the increased number of MDPs, does not increase indefinitely with the number of training levels/scenes. Instead, we observe that the sample variance starts hitting a plateau after around $1000$ training levels (see Fig. \ref{fig:variance_saturation}). Interestingly, \cite{cobbe2019leveraging} also report that, for most ProcGen environments, the gap between train and test performance starts reducing around the same range of training levels.

These observations support the intuition that there is a direct correlation between the equivalent sample variance $\mathbf{E}_{s,a,\mathcal{M}}[\psi^2(s,a,\mathcal{M})]$ and the diversity of experiences required for competitive generalization \footnote{To be precise, the statement strictly holds when the sample variance $\mathbf{E}_{s,a,\mathcal{M}}[\psi^2(s,a,\mathcal{M})]$ is computed while training using the same network architecture (e.g. IMPALA-CNN, \cite{espeholt2018impala}) and algorithm choice (e.g. baseline PPO).}. Thus, for ProcGen environments, most of the variance in the game scenarios is already well captured in the first $500 \sim 1000$ levels and adding further training data does not change the statistics of underlying sample $(\psi(s,a,\mathcal{M}))$ distribution by a significant amount.
We still do not completely understand the slight drop in variance of $\psi(s,a,\mathcal{M})$ samples at the end of the curve, and believe that further research is necessary to explain this phenomenon.

\section{Analysing Gains from the Dynamic Model}
The effectiveness of the dynamic model in reducing sample variance, though theoretically sound, is quite difficult to demonstrate in practice. This is because as the dynamic agent learns a better policy, it starts exploring more areas from the game scenes and thus is coincident with an accompanying increase in the sample variance. 

\begin{figure}[ht]
\vskip 0.2in
\begin{center}
\centerline{\includegraphics[width=\textwidth]{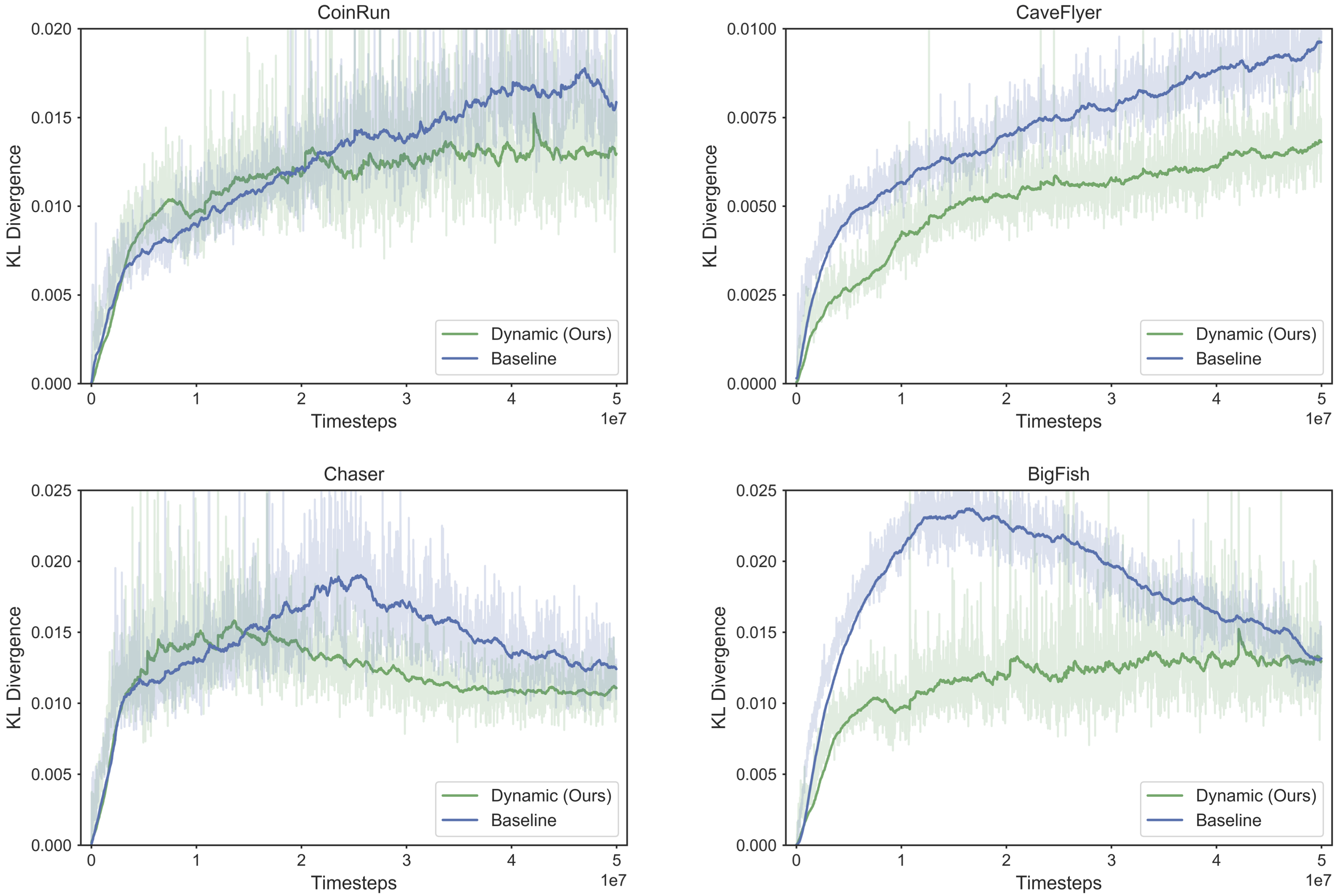}}
\caption{Dynamic model learns more robust policy updates with lower KL divergence while achieving better final performance. Refer to the performance comparison between baseline and dynamic models in Table 1 and Fig. 4 from the original paper.}
\label{fig:kl_divergence_curves}
\end{center}
\vskip -0.2in
\end{figure}

We however observe that the dynamic model leads to a lower KL divergence between the new and old policies for the PPO \cite{schulman2017proximal} algorithm. Results are shown in Fig. \ref{fig:kl_divergence_curves}. We clearly see that the dynamic model consistently achieves lower KL divergence than the baseline models while resulting in a better overall episode reward. This mixture of lower KL divergence and better performance indicate that the dynamic model is more stable than its baseline counterpart. Figuratively, this implies that the dynamic model learns to make much smaller but meaningful steps towards the optimal policy \cite{schulman2015trust}.

\end{document}